\documentclass[
]{ceurart}

\usepackage{listings}
\usepackage{inputenc}
\usepackage{multirow}
\usepackage{comment}
\usepackage[normalem]{ulem}
\useunder{\uline}{\ul}{}
\usepackage{todonotes}
\usepackage{booktabs}
\usepackage{enumitem}

\begin{document}

\copyrightyear{2022}
\copyrightclause{Copyright for this paper by its authors.
  Use permitted under Creative Commons License Attribution 4.0
  International (CC BY 4.0).}

\conference{MediaEval'22: Multimedia Evaluation Workshop,
  January 13--15, 2023, Bergen, Norway and Online}

\title{Managing Large Dataset Gaps in Urban Air Quality Prediction:  DCU-Insight-AQ at MediaEval 2022}


\author[1,2]{Dinh Viet Cuong}[%
    email=dinh.cuong2@mail.dcu.ie,
]

\author[2]{Phuc H. Le-Khac}[%
    orcid=0000-0002-0504-5844,
    email=khac.le2@mail.dcu.ie,
]

\author[2]{Adam Stapleton}[%
    orcid=0000-0003-1233-211X,
    email=adam.stapleton9@mail.dcu.ie,
]

\author[3]{Elke Eichlemann}[%
    orcid=0000-0001-9516-7951,
    email=elke.eichelmann@ucd.ie,
]

\author[1,2]{Mark Roantree}[%
    orcid=0000-0002-1329-2570,
    email=mark.roantree@dcu.ie]

\author[1,2]{Alan F. Smeaton}[%
    orcid=0000-0003-1028-8389,
    email=alan.smeaton@dcu.ie
]
\cormark[1]

\address[1]{Insight Centre for Data Analytics, Dublin City University, Ireland}
\address[2]{School of Computing, Dublin City University, Ireland}
\address[3]{School of Biology and Environmental Science, University College Dublin, Ireland}

\cortext[1]{Corresponding author.}

\begin{abstract}
Calculating an Air Quality Index (AQI) typically uses data streams  from air quality sensors deployed at fixed locations and the calculation is a real time process.  If one or a number of sensors are broken or offline, then the real time AQI value cannot be computed.   Estimating AQI values for some point in the future is a predictive process and uses historical AQI values to train and build models.  In this work we focus on gap filling in air quality data where the task is to predict the AQI at 1, 5 and 7 days into the future. The scenario is where one or a number of air, weather and traffic sensors are offline and explores prediction accuracy under such situations. The work is part of the MediaEval'2022 Urban Air: Urban Life and Air Pollution task submitted by the DCU-Insight-AQ team and uses multimodal and crossmodal data consisting of AQI, weather and CCTV traffic images for air pollution prediction.

\end{abstract}

\maketitle

\section{Introduction}
\label{sec:intro}

The Urban Life and Air Pollution task at MediaEval 2022 required participants to predict the air quality index (AQI) value at +1, +5 and +7 days using an archive of air quality, weather and images from 16 CCTV cameras, one image taken every 60 seconds  \cite{UA22}. Participating groups were required to download the data from online sources for local processing.
Gaps in air quality datasets are common with the problem exacerbated for data gathered in poorer or developing countries \cite{PINDER2019116794, Falge2001, Hui2004, Moffat2007, Kim2020}. In this paper we describe how we addressed the very large gaps in data that we encountered in the data we downloaded.


\section{Methodology}
\label{sec:approach}
The biggest issue in this research challenge was the significant  gaps in the training data, highlighted in Figure~\ref{fig:availability} for the data we downloaded, and regarded as a common issue with climate datasets.
Because participants downloaded data independently, and because data servers had different periods of downtime, it is likely that participants have different, but overlapping training data and perhaps  other participants managed to download more data than we did. Even so,  the amount of data we downloaded  allows us to focus on the challenge of data gaps.
This is directly addressed in our research methodology by first identifying the sensitivity of the data gaps and adopting a counter-measure to eliminate gap data. Our method comprises 4 steps:

\begin{itemize}[leftmargin=*]
    \item Step 1. Data Analysis. This performs a statistical summary of datasets including the computation of spatial data related to the locations of air quality stations and cameras.
    \item Step 2. Gap Filling. Elimination or maximising the reduction in the gaps in air quality data.
    \item Step 3. Processing CCTV camera images. This step transforms each image into a set of features that can be combined with the air quality feature set.
    \item Step 4. Model Building. This step builds an experimental platform using different machine learning model configurations together with different feature sets to identify the best performing model/feature set combination.  
\end{itemize}
In the remainder of this section, we described the first 3 steps in detail and in Section~\ref{sec:exper}, we describe our approach to model building for the air quality prediction task.


\subsection{Data Analysis}

\begin{figure}[ht!]
    \centering
    \includegraphics[width=1.0\textwidth]{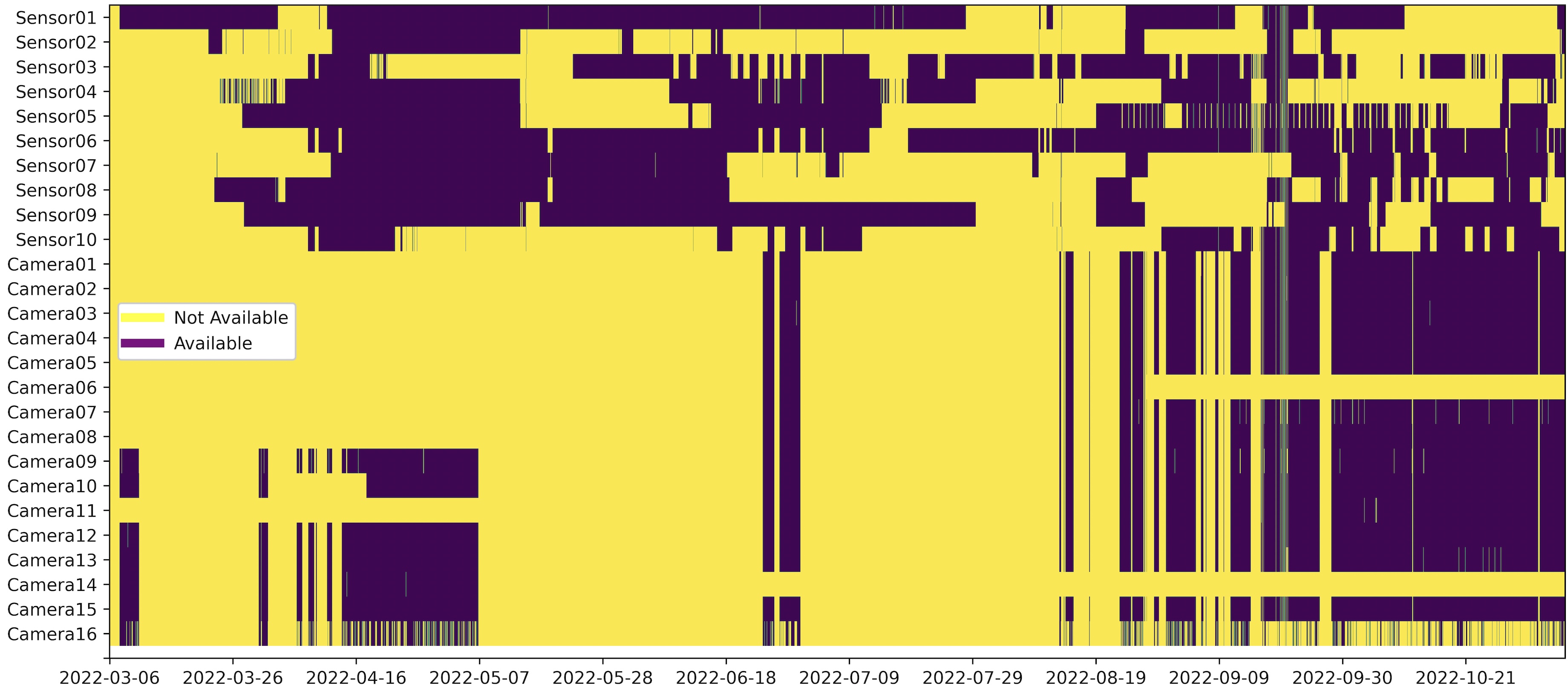}
    \caption{Missing data (shown in yellow) from across all 10 air quality measurement stations and 16 CCTV cameras for an 8-month period, early March to early November 2022.}
    \label{fig:availability}
\end{figure}

The downloaded air quality data are collected at 10 monitoring stations in Dalat City, Vietnam from March 2020  to 7th Nov 2022. The data includes air pollutant concentration for NO$_2$(ppm), CO, SO$_2$, O$_3$, PM1.0, PM2.5, PM10 as well as environmental measures namely temperature, humidity, UV, rainfall. In addition, traffic data, in form of images, was recorded every  minute from 16 CCTV cameras across Dalat City. Figure~\ref{fig:availability} shows the availability of the  dataset as downloaded by our group. This shows huge gaps in data availability.  In our model building we use the first 80\% of available data for training machine learning models and the remaining 20\% for validation.

\subsection{Image Processing}
For the CCTV data, we downloaded a total of  398,412 images from across all 16 cameras, which took approximately 215GB of storage. 
If all data had been available and  downloaded, there would be approximately 16 cameras x 8 months x 30 days x 24 hours x 60 minutes = 5,529,600 images  so our download represents 7.2\% of the theoretical maximum. 

We  re-sized each image to 640x640 and processed each using  a medium-sized YOLOv6 object-detector \cite{yolov6} pre-trained on the COCO dataset \cite{coco}. This performs well with a balanced trade-off between speed and accuracy.
From an output with more than 80 object categories, we used the average of 4 vehicle types as a proxy for traffic volume with an average detection per image of 2.58 (cars), 3.90 (motorcycles), 0.16 (bus) and 0.25 (trucks). 
These values are used directly as features for our predictive model without  further post-processing.

\subsection{Gap Filling}
Our approach to gap filling  used 3 different feature sets. 

\begin{itemize}[leftmargin=*]
    \item Rolling Window (FS1). We generate training data using rolling windows sliding hourly over the data. 
    Following some experimentation, we determine the historical data length to input to the model (size of window) to be 2 days. To address missing data, we remove windows with big gaps and fill windows with small gaps of less than 30\%  or have data from more than 4 stations. 
    \item Rolling Window with Gap Filling (FS2).  Spatial interpolation using a tree-based gradient boosting model \cite{lightgbm} was used to infer missing values. A separate model to predict each of the air pollutants at each station was trained using all available data from  other stations at that timestamp.  The LightGBM model carries the same sparsity-aware learning methods as XGBoost \cite{xgboost} while improving on efficiency. Such methods for handling sparse arrays allow a model to learn from incomplete data. 
    \item Rolling Window with Gap Filling and Image/Traffic Features (FS3).  This feature set uses the output from the count of the numbers of cars, trucks, motorcycles and buses detected in the CCTV images from the 16 cameras.

\end{itemize}

\noindent
We convert all pollutant concentration values into AQI values according to the formula described in \cite{AQIformula}. 
Air quality, environmental  and CCTV features are aggregated into hourly mean values which are normalised to a mean of 0.0 and standard deviation of 1.0 before  input into  a selection of different machine learning models.

\section{Experiments}
\label{sec:exper}
\subsection{Experimental Approach}

We used baseline machine learning models including multi-layer perceptrons, long short-term memory (LSTM) \cite{LSTM} and LSTM-GNN \cite{LSTMGNN} where the prediction is treated as timeseries forecasting. We also used a spatio-temporal graph neural network \cite{LSTMGNN} and our own  neural network architecture labelled TemAtt and SpaTemAtt, designed to capture more complex temporal and spatio-temporal patterns.


\subsection{Results}

\begin{table}[h!]
\begin{tabular}{lllll|lll}
\hline
\multirow{2}{*}{Feature Set} & \multirow{2}{*}{Model} & \multicolumn{3}{l|}{Station (recent data)}                                                 & \multicolumn{3}{l}{Station (no recent data)}                                                                    \\ \cline{3-8} 
                            &                        & \multicolumn{1}{l|}{+1day}          & \multicolumn{1}{l|}{+5days}         & +7days         & \multicolumn{1}{l|}{+1day}          & \multicolumn{1}{l|}{+5days}         & \multicolumn{1}{l}{+7days}         \\ \hline
FS1                         & MLP                    & \multicolumn{1}{l|}{20.51}          & \multicolumn{1}{l|}{27.77}          & 26.75          & \multicolumn{1}{l|}{45.46}          & \multicolumn{1}{l|}{48.75}          & \multicolumn{1}{l}{46.06}          \\
FS1 (run 1)                         & LSTM                   & \multicolumn{1}{l|}{19.94}          & \multicolumn{1}{l|}{25.84}          & {\ul 25.42}    & \multicolumn{1}{l|}{44.15}          & \multicolumn{1}{l|}{46.87}          & \multicolumn{1}{l}{46.38}          \\
FS2  (run 3)                      & LSTM                   & \multicolumn{1}{l|}{{\ul 18.94}}    & \multicolumn{1}{l|}{25.52}          & \textbf{24.95} & \multicolumn{1}{l|}{43.98}          & \multicolumn{1}{l|}{46.66}          & \multicolumn{1}{l}{45.36}          \\
FS2  (run 4)                       & TemAtt                 & \multicolumn{1}{l|}{\textbf{18.57}} & \multicolumn{1}{l|}{{\ul 24.96}}    & 27.73          & \multicolumn{1}{l|}{44.36}          & \multicolumn{1}{l|}{46.60}          & \multicolumn{1}{l}{48.67}          \\
FS2                         & LSTM-GNN               & \multicolumn{1}{l|}{23.84}          & \multicolumn{1}{l|}{25.99}          & 30.48          & \multicolumn{1}{l|}{39.84}          & \multicolumn{1}{l|}{40.21}          & \multicolumn{1}{l}{44.48}          \\
FS2  (run 2)                       & SpaTemAtt              & \multicolumn{1}{l|}{19.93}          & \multicolumn{1}{l|}{\textbf{24.91}} & 29.92          & \multicolumn{1}{l|}{\textbf{30.93}} & \multicolumn{1}{l|}{\textbf{34.42}} & \multicolumn{1}{l}{\textbf{34.85}} \\
FS3   (run 5)                      & SpaTemAtt              & \multicolumn{1}{l|}{20.43}          & \multicolumn{1}{l|}{25.18}          & 31.75          & \multicolumn{1}{l|}{{\ul 35.21}}    & \multicolumn{1}{l|}{{\ul 38.46}}    & \multicolumn{1}{l}{{\ul 40.32}}    \\ \hline
\end{tabular}
\caption{Best Performing Model Configurations}
\label{tab:results}
\end{table}

Table \ref{tab:results} presents results after our own validation using bold font to highlight the best performing model with the second best results underlined.  The first 2 columns show the features and models used where runs marked 1-5 represent our 5 submitted prediction attempts. The remaining 6 columns, which present RMSE scores, are split according to those stations that continued to provide data, ``Station (recent data)'', and those that stopped supplying data due to malfunctions, ``Station (no recent data)'' at the station. For the latter, it was necessary to make predictions for these stations based on recent data at other stations. For both station categories, we provide our  accuracy for 1-day ahead, 5-days ahead and 7-days ahead predictions. Validation uses root mean squared error, so the lower the error cost, the better the model. 

\textbf{Analysis.} Stations with recent data far out-performed those without so we will focus our discussion on the former. As expected, the predictive accuracy for all models decreases as the forecast window increased, significantly for 5 and 7 day predictions.
Feature set 2 (FS2) was the best performing  across all time intervals. This highlighted the benefit of our gap filling method but also indicated that the camera data did not deliver any added performance.  
In terms of models, no single model performed best  although our Temporal Graph model (TemAtt) was the best-performing model for 1-day predictions and the Spatio-Temporal Graph model (SpaTempAtt) was best performing for 5-day predictions. Finally, all test models, including both of our Graph models, showed significant degradation in performance as sensitivity decreases.

\section{Conclusions and Lessons Learned}

Some high-level conclusions from our results show that our gap filled data have a positive impact on the predictions, it approximates missing values better and adds spatial information from other locations.
Time series specialised models are better at predictions in short-term as LSTM and TemAtt architectures out-perform MLP and simpler models marginally.

While the results provide answers to some questions, there are may other unanswered ones.  For example it is not clear why  the TemAtt model outperformed the SpaTemAtt model. It may be because the inclusion of spatial info made SpaTemAtt weaker somehow. We were surprised by the comparatively poor performance of the LSTM-GNN.  For us, the CCTV images added no value as FS3 featured outside the top-2 across all prediction intervals possibly because there is little overlap between CCTV and air quality data thus little for the models to train on. One other reason for this could be that additional traffic features do not really provide new information. Even if we assume air quality is affected by traffic, was our vehicle counting too naive or is there another explanation.



\subsection*{Acknowledgements}

This work was supported in part by  Science Foundation Ireland through the the Insight Centre for Data Analytics (SFI/12/RC/2289\_P2) and the Centre for Research Training in Machine Learning (18/CRT/6183). We thank the organisers for running the task.

\def\bibfont{\small} 
\bibliography{references} 

\end{document}